\documentclass[aps,pre,twocolumn,superscriptaddress,amsmath,amssymb,floatfix,longbibliography]{revtex4-2}

\usepackage{graphicx}
\usepackage{bm}
\usepackage{booktabs}
\usepackage{microtype}
\usepackage[hidelinks]{hyperref}
\newcommand{\dd}{\mathrm{d}}
\newcommand{\simplex}{\Delta^{N-1}}
\newcommand{\Abe}{\mathrm{A}}

\begin{document}

\title{Entropy-Generated Attention Beyond Softmax and Entmax: Kaniadakis and Reciprocal-Symmetric Abe Operators}

\author{Gunn Kim}
\email{gunnkim@sejong.ac.kr}
\affiliation{Department of Physics, Sejong University, Seoul 05006, Republic of Korea}

\date{\today}

\begin{abstract}
We derive two attention operators from generalized statistical entropies.  Kaniadakis entropy yields an exact full-support normalization whose weights and low-score sensitivities decay algebraically, rather than exponentially as in Softmax or by exact truncation as in entmax.  Classical Abe entropy yields an implicit reciprocal-symmetric operator.  With $q=e^\epsilon$, the involution $q\leftrightarrow q^{-1}$ removes every odd correction about Softmax; we obtain the normalized second- and fourth-order terms, including the deformation of the normalization multiplier.  These stationary laws follow from a Fisher-metric Lagrangian on the probability simplex, whose Shannon sector recovers scaled dot-product Softmax.  We also give a tangent-gradient test for deciding whether changing the entropy changes the attention profile or only its scale.  R\'enyi and two-parameter Sharma--Mittal entropies retain the Tsallis--entmax inverse-gradient shape, but their global moments make the effective temperature input dependent when the external temperature is fixed.  Distinguishing profile-shape equivalence from fixed-parameter operator equivalence separates new normalization shapes from adaptive rescalings and organizes the operators by support, tail behavior, and realization complexity.
\end{abstract}

\maketitle

\section{Introduction}

Analytical mechanics replaces the search for equations of motion by the search for a scalar functional.  Once an action $\mathcal A[q]$ is specified, its stationary variation selects the dynamics~\cite{goldstein2002},
\begin{equation}
 \delta\mathcal A[q]=0
 \quad\Longrightarrow\quad
 \frac{\dd}{\dd t}\frac{\partial L}{\partial\dot q}
 -\frac{\partial L}{\partial q}=0.
 \label{eq:EL-intro}
\end{equation}
The same logic reappears in statistical mechanics, where competition between energy and entropy selects a stationary distribution~\cite{jaynes1957}.  In both settings the functional precedes the law.  We ask what attention laws are selected when the scalar functional is changed.

Modern neural attention emerged as a trainable alignment mechanism~\cite{bahdanau2014} and became the central operation of the Transformer~\cite{vaswani2017}.  For a fixed query $\bm q$, scaled dot-product attention assigns
\begin{equation}
 \rho_i=\frac{\exp(z_i/\tau)}{\sum_j\exp(z_j/\tau)},
 \qquad z_i=\bm q\!\cdot\!\bm k_i,
 \qquad \tau=\sqrt{d_k},
 \label{eq:standard}
\end{equation}
and returns $\sum_i\rho_i\bm v_i$.  The factor $1/\sqrt{d_k}$ controls the score variance and limits premature saturation~\cite{vaswani2017}.  Softmax is also the maximizer of a linear score functional regularized by Shannon entropy~\cite{shannon1948,bridle1990}.  More generally, regularized prediction maps generate probability-valued attention rules from convex or entropy regularizers~\cite{niculae2017,blondel2019}; sparsemax is a Euclidean simplex projection~\cite{martins2016}, and $\alpha$-entmax follows from Tsallis regularization~\cite{peters2019,tsallis1988}.  Thus entropy already generates useful attention laws.  The question here is which other statistical entropies generate genuinely different ones.

The answer is not one entropy--one operator.  R\'enyi~\cite{renyi1961} and Sharma--Mittal~\cite{sharma1975} entropies preserve the Tsallis inverse-gradient shape, even though their global moments change the relation between an external temperature and the realized profile.  Monotone transforms of the same power functional are also known to share maximum-entropy optimizers under linear constraints~\cite{hino2026}.  We make the attention-specific distinction between equivalence of stationary profile shape and equality of the score-to-weight map at fixed parameters.  A tangent-gradient test then identifies two escapes: Kaniadakis entropy~\cite{kaniadakis2002,kaniadakis2005} gives an exact dense operator with algebraic tails and an algebraic low-score Jacobian, while classical Abe entropy~\cite{abe1997} gives a reciprocal-symmetric operator whose expansion about Softmax contains only even powers.  To our knowledge, these two laws have not previously been developed as attention normalizations.

We embed the stationary construction in analytical mechanics.  The probability simplex carries the Fisher--Rao metric~\cite{fisher1925,amari2000}; it fixes the kinetic term, and the entropy-dependent free energy supplies the potential.  Variational free-energy methods are established in machine learning and theoretical neuroscience~\cite{dayan1995,friston2005,friston2010}, but their optimized densities represent latent causes under generative models.  Here $\bm\rho$ is directly the attention distribution over keys, the scores and simplex constraint are held fixed, and the entropy is varied.  The full Euler--Lagrange equation specifies the broader dynamical structure, while all new operators in this paper are derived in its stationary sector.  Recent uses of Fisher geometry in attention address representation or Bayesian geometry~\cite{bathula2026,rubacher2026}, rather than this entropy-to-operator construction.

Complementary variational views treat attention as one-sided entropic transport~\cite{litman2025}, impose doubly stochastic Sinkhorn structure~\cite{sander2022}, learn transport priors~\cite{litman2026goat}, modify a looped Transformer's training landscape~\cite{lam2026}, or introduce pairwise mean-field couplings~\cite{wang2026via}.  A recent survey reviews broader uses of Tsallis statistics in artificial intelligence~\cite{dacosta2026}.  Here the scores and simplex constraint remain fixed so that changing the entropy isolates the normalization law.

The logical structure is
\begin{equation}
 \begin{gathered}
 \text{generalized entropy}\rightarrow
 \text{stationary attention law}\\[-2pt]
 \rightarrow\text{tangent-gradient test}\\[-2pt]
 \rightarrow\text{new operator or reparameterization}.
 \end{gathered}
 \label{eq:logic}
\end{equation}
The first arrow is the stationary counterpart of Eq.~\eqref{eq:EL-intro}: a chosen functional generates a law.  The remaining arrows expose the non-injectivity created by normalization.  We therefore separate stationary profile shape, the fixed-parameter score-to-weight map, and the numerical procedure required to realize it.

\section{Variational map}
\label{sec:stationary}

\subsection{Fisher--Lagrangian construction}

For a single query, let $\bm\rho=(\rho_1,\ldots,\rho_N)$ lie on the probability simplex $\simplex$, and define score energies by
\begin{equation}
 E_i=-z_i=-\bm q\!\cdot\!\bm k_i.
 \label{eq:energy}
\end{equation}
For a concave entropy $S(\bm\rho)$, we consider
\begin{equation}
 \mathcal F_S(\bm\rho)
 =\sum_i\rho_iE_i-T S(\bm\rho),
 \qquad T>0,
 \label{eq:potential}
\end{equation}
under $\sum_i\rho_i=1$ and $\rho_i\geq0$.  Here $T$ fixes the scale of entropy regularization relative to the scores.  It is an operator parameter, not a claim of coupling to a physical heat bath.

The Fisher--Rao line element in the simplex interior is
\begin{equation}
 \dd s_{\rm F}^{2}=\sum_i\frac{(\dd\rho_i)^2}{\rho_i},
 \qquad \sum_i\dd\rho_i=0.
 \label{eq:fisher-metric}
\end{equation}
The amplitude coordinates $x_i=2\sqrt{\rho_i}$ map the simplex to the positive orthant of the sphere $\sum_i x_i^2=4$ and give $\sum_i(\dd x_i)^2=\dd s_{\rm F}^2$.  This fixes the kinetic term without introducing an additional coordinate-dependent metric ansatz.  With an effective information inertia $m_{\rm att}>0$, we define
\begin{equation}
 \mathcal L_S
 =\frac{m_{\rm att}}{2}\sum_i\frac{\dot\rho_i^2}{\rho_i}
 -\mathcal F_S(\bm\rho)
 +\lambda(t)\left(\sum_i\rho_i-1\right).
 \label{eq:fisher-lagrangian}
\end{equation}
We interpret $t$ as a continuous layer-depth coordinate, following the dynamical-systems and Neural-ODE view in which residual-layer updates discretize a continuous flow~\cite{weinan2017,chen2018node}.  Thus $\bm\rho(t)$ denotes the attention distribution along depth, and a discrete stack samples this trajectory at successive layers.  In the nonstationary equation the score potential may more generally be depth dependent, $E_i=E_i(t)$; the stationary operators studied here hold the scores fixed and set $\dot{\bm\rho}=\ddot{\bm\rho}=0$.

Hamilton's principle applied to Eq.~\eqref{eq:fisher-lagrangian} gives
\begin{equation}
 m_{\rm att}\left(
 \frac{\ddot\rho_i}{\rho_i}
 -\frac{1}{2}\frac{\dot\rho_i^2}{\rho_i^2}
 \right)
 +E_i-T\frac{\partial S}{\partial\rho_i}
 =\lambda(t).
 \label{eq:euler-lagrange-attention}
\end{equation}
Equation~\eqref{eq:euler-lagrange-attention} is the entropy-dependent Euler--Lagrange equation associated with the chosen potential.  In the stationary sector, $\dot\rho_i=\ddot\rho_i=0$, it reduces to
\begin{equation}
 E_i+\mu-T\frac{\partial S}{\partial\rho_i}=0,
 \qquad \sum_i\rho_i=1,
 \label{eq:master}
\end{equation}
where $\mu=-\lambda$ is a scalar normalization multiplier.  All attention operators analyzed below are obtained in this static sector and therefore depend only on $\mathcal F_S$ and the simplex constraint, not on the specific form or overall normalization of the kinetic term.  We retain the full Euler--Lagrange equation to make explicit that the stationary variational problem is one sector of a geometrically specified continuous-depth structure.  Whether attention paths in trained networks approximately extremize this action is a separate empirical question.  Equivalently,
\begin{equation}
 \bm\rho^*=\arg\min_{\bm\rho\in\simplex}
 \left[\sum_i\rho_iE_i-T S(\bm\rho)\right].
 \label{eq:min}
\end{equation}
For strictly concave $S$, the objective is strictly convex and the optimizer is unique.  Equation~\eqref{eq:master} is the master equation of the paper.  It shows that the inverse entropy gradient maps scores to probabilities.  All equivalence statements below are restricted to parameter ranges in which the relevant entropy is concave and the stationary branch under discussion is well defined.

We use the linear mean-energy term $\sum_i\rho_iE_i$.  Generalized thermostatistics also uses escort constraints proportional to $\rho_i^q$, which lead to different stationary conditions away from the Shannon limit~\cite{tsallisMP1998}; related transformations and tail constraints are discussed in Ref.~\cite{bercher2008}.  The linear convention is especially natural for attention because the output is a linear weighted sum of values.  Claims below are conditional on this choice.

\subsection{Profile-shape and fixed-parameter equivalence}

Only changes of the entropy gradient along the simplex can alter stationary score differences.  A component-independent term in $\nabla S$ is absorbed into $\mu$, while a positive scale common to all components can be absorbed into an effective temperature. Let us define the Euclidean projector onto the tangent space of the simplex by
\begin{equation}
 \Pi_\Delta
 =I-\frac{1}{N}\bm 1\bm 1^{\mathsf T},
 \label{eq:tangent-projector}
\end{equation}
where \(I\) is the \(N\times N\) identity matrix and
\(\bm 1=(1,\ldots,1)^{\mathsf T}\) is the all-ones column vector.
The tangent space consists of variations satisfying
\(\sum_i\delta\rho_i=0\).
Two entropies have the same stationary \emph{profile shape} whenever
\begin{equation}
 \Pi_\Delta\nabla S_1(\bm\rho)
 =a(\bm\rho)\Pi_\Delta\nabla S_2(\bm\rho),
 \qquad a(\bm\rho)>0.
 \label{eq:equivalence}
\end{equation}
Equivalently, their full gradients may differ as
\begin{equation}
 \nabla S_1=a(\bm\rho)\nabla S_2+b(\bm\rho)\bm 1.
 \label{eq:gauge}
\end{equation}
The $b$ term is a normalization gauge and the $a$ term changes the score-to-entropy scale.  Indeed, projecting Eq.~\eqref{eq:master} gives $\Pi_\Delta\bm z=-T\Pi_\Delta\nabla S$.  Hence the same profile $\bm\rho$ is stationary for the same score differences after the replacement $T_2=T_1a(\bm\rho)$.  This is \emph{profile-shape equivalence}.  It is weaker than equality of the machine-learning operators at fixed external parameters.  A single temperature reparameterization works for every score vector only when $a$ is independent of $\bm\rho$; we call that \emph{fixed-parameter equivalence}.  When $a$ depends on a moment of the output, the two maps share a profile shape but one realizes it through an input-dependent effective temperature.

For component-separable entropies, the test has a simple scalar interpretation.  If $\partial S_m/\partial\rho_i=h_m(\rho_i)$ and Eq.~\eqref{eq:equivalence} holds throughout an open part of a simplex with $N\geq4$, differences among four independently variable components force $h_1(x)=A h_2(x)+B$ on the corresponding interval when fixed-parameter equivalence is claimed.  The constant $B$ is the normalization gauge.  Failure of this affine relation proves that the inverse-gradient shapes differ.

Equation~\eqref{eq:equivalence} also gives a quick numerical falsification test.  At a generic nonuniform interior point for which $\|\bm u\|_2\|\bm v\|_2>0$, set $\bm u=\Pi_\Delta\nabla S_1$ and $\bm v=\Pi_\Delta\nabla S_2$, and compute
\begin{equation}
 a_*=\frac{\bm u\!\cdot\!\bm v}{\bm v\!\cdot\!\bm v},
 \qquad
 R=\frac{\|\bm u-a_*\bm v\|_2}{\|\bm u\|_2}.
 \label{eq:residual-test}
\end{equation}
If $a_*\leq0$ or $R>0$ beyond numerical tolerance at even one such point, the two entropies do not share a stationary profile shape for those parameter values.  Conversely, $R=0$ at one sampled point is only diagnostic; equivalence requires the functional identity to hold throughout the relevant simplex domain.  Raw ratios of gradient components are not a gauge-invariant substitute because the additive term $b(\bm\rho)\bm 1$ in Eq.~\eqref{eq:gauge} changes those ratios.

For a binary distribution the tangent space is one-dimensional, so a direction-only test is necessarily degenerate.  The operator distinctions below concern ordinary attention vectors with at least three, and generically many, components; the non-equivalence argument for arbitrary component values is stated for $N\geq4$.

The \emph{realization complexity} separately records whether a specified map is obtained by a closed-form sum, an active-set threshold, a scalar normalization root, or a self-consistent global moment.  Two entropies can share a stationary profile shape while defining different fixed-$T$ maps and requiring different numerical procedures.

Three further properties complete the comparison.  A divergent entropy gradient at $\rho_i\to0^+$ generally enforces full support, whereas a finite gradient permits a boundary optimum.  Componentwise algebraic invertibility determines whether only a normalization root remains.  Finally, the asymptotic inverse gradient determines exponential, algebraic, inverse-linear, or cutoff behavior.  These are structural regimes encountered among the entropy families studied here, not a claim of an exhaustive universal taxonomy.

\section{Explicit entropy-generated operators}
\label{sec:explicit}

\subsection{Shannon entropy and Softmax}

For Shannon entropy~\cite{shannon1948}
\begin{equation}
 S_{\rm Sh}(\bm\rho)=-\sum_i\rho_i\ln\rho_i,
\end{equation}
Eq.~\eqref{eq:master} becomes
\begin{equation}
 E_i+\mu+T(\ln\rho_i+1)=0.
\end{equation}
Consequently,
\begin{equation}
 \rho_i^{\rm Sh}
 =\frac{e^{-E_i/T}}{\sum_j e^{-E_j/T}}
 =\frac{e^{z_i/T}}{\sum_j e^{z_j/T}}.
 \label{eq:softmax}
\end{equation}
With $T=\sqrt{d_k}$, Eq.~\eqref{eq:softmax} is Eq.~\eqref{eq:standard}.  The logarithmic divergence of $\partial S_{\rm Sh}/\partial\rho_i$ at the boundary guarantees positive weights for finite scores.  An energy increase suppresses a weight exponentially.  This is the canonical baseline against which the generalized cases should be compared, rather than evidence that a network is literally in thermal equilibrium.

\subsection{Tsallis entropy and entmax}

Tsallis entropy is~\cite{tsallis1988}
\begin{equation}
 S_q(\bm\rho)=\frac{1-\sum_i\rho_i^q}{q-1},
 \qquad q>1.
 \label{eq:tsallis-entropy}
\end{equation}
Its derivative is $-q\rho_i^{q-1}/(q-1)$.  Because the derivative remains finite at $\rho_i=0$ for $q>1$, the optimizer may lie on the boundary.  Applying the Karush--Kuhn--Tucker conditions gives
\begin{equation}
 \rho_i^{(q)}
 =\left[\frac{q-1}{qT}(z_i-\theta)\right]_+^{1/(q-1)},
 \label{eq:tsallis-map}
\end{equation}
where $[u]_+=\max(u,0)$ and $\theta$ is the unique threshold satisfying $\sum_i\rho_i^{(q)}=1$.

Equation~\eqref{eq:tsallis-map} is the Tsallis-regularized prediction map used in the $\alpha$-entmax family~\cite{peters2019}, up to entropy and temperature conventions.  The limit $q\to1$ recovers Softmax.  For $q=2$, the exponent is one and the map becomes a Euclidean simplex projection equivalent to sparsemax after rescaling~\cite{martins2016}.  Thus the same stationary principle connects smooth dense attention to exact sparsity.

It is important not to insert an external partition function after Eq.~\eqref{eq:tsallis-map}.  The threshold $\theta$ already enforces normalization, and a separate division would generally change the solution because a positive-part power does not have the ordinary exponential shift property.  Likewise, an unnormalized canonical $q$-exponential written with a fixed reference energy is not a second stationary family under the same objective; differences usually reflect constraint or parameter conventions~\cite{tsallisMP1998,bercher2008}.

The support
\begin{equation}
 \mathcal I_q=\{i:z_i>\theta\}
\end{equation}
is selected jointly with the weights.  Lowering $T$ or increasing $q$ typically reduces the active set, although the exact dependence is score-specific.  Sparsity is therefore a consequence of entropy geometry, not a post-processing truncation.

\subsection{Kaniadakis entropy and algebraic-tail attention}

Kaniadakis entropy~\cite{kaniadakis2002,kaniadakis2005} is
\begin{equation}
 S_\kappa(\bm\rho)
 =-\sum_i\rho_i
 \frac{\rho_i^\kappa-\rho_i^{-\kappa}}{2\kappa},
 \qquad 0<|\kappa|<1,
 \label{eq:kentropy}
\end{equation}
with the Shannon value defined by the continuous limit $\kappa\to0$.  It is even under $\kappa\mapsto-\kappa$.  Its first derivative is
\begin{equation}
 \frac{\partial S_\kappa}{\partial\rho_i}
 =-\frac{(1+\kappa)\rho_i^\kappa
 -(1-\kappa)\rho_i^{-\kappa}}{2\kappa}.
 \label{eq:kderivative}
\end{equation}
and strict concavity follows from
\begin{equation}
 \frac{\partial^2 S_\kappa}{\partial\rho_i^2}
 =-\frac{(1+\kappa)\rho_i^{\kappa-1}
 +(1-\kappa)\rho_i^{-\kappa-1}}{2}<0.
 \label{eq:kconcavity}
\end{equation}
Let $x_i=(z_i-\theta)/T$ and $u_i=\rho_i^\kappa$.  The stationary equation is
\begin{equation}
 (1+\kappa)u_i^2-2\kappa x_i u_i-(1-\kappa)=0.
\end{equation}
The positive root yields the exact linear-constraint solution
\begin{equation}
 \rho_i^{(\kappa)}=
 \left[
 \frac{\kappa x_i+\sqrt{1-\kappa^2+\kappa^2x_i^2}}
 {1+\kappa}
 \right]^{1/\kappa}.
 \label{eq:kmap}
\end{equation}
The scalar $\theta$ is fixed by $\sum_i\rho_i^{(\kappa)}=1$.  Each component is continuous and strictly decreasing in $\theta$, and the sum ranges from infinity to zero as $\theta$ runs from $-\infty$ to $+\infty$; the normalization root therefore exists and is unique.

Equation~\eqref{eq:kmap} has the exact conventional $\kappa$-exponential form
\begin{equation}
 \exp_\kappa(x)=\left(\sqrt{1+\kappa^2x^2}+\kappa x\right)^{1/\kappa},
\end{equation}
namely
\begin{equation}
 \begin{split}
 \rho_i^{(\kappa)}&=\alpha_\kappa
 \exp_\kappa\!\left(\frac{x_i}{\lambda_\kappa}\right),\\
 \alpha_\kappa&=\left(\frac{1-\kappa}{1+\kappa}\right)^{1/(2\kappa)},
 \qquad \lambda_\kappa=\sqrt{1-\kappa^2}.
 \end{split}
 \label{eq:kexp-form}
\end{equation}
Both constants are fixed by differentiating the trace-form entropy; they are not an independently chosen partition factor or temperature.  The scalar root for $\theta$ supplies the actual normalization.

The tangent-gradient criterion also shows directly that Eq.~\eqref{eq:kmap} is not a temperature-reparameterized Tsallis operator.  For $N\geq4$, equivalence throughout the simplex would require the scalar functions in the two entropy gradients to be affinely related,
\begin{equation}
 -\frac{(1+\kappa)x^\kappa-(1-\kappa)x^{-\kappa}}
 {2\kappa}
 =A x^{q-1}+B
 \label{eq:kappa-escape}
\end{equation}
for all $x\in(0,1)$, where $A$ and $B$ represent temperature rescaling and normalization gauge.  For $0<|\kappa|<1$, the left-hand side contains two independent nonzero powers, $x^\kappa$ and $x^{-\kappa}$, whereas the Tsallis gradient contains only one.  No constants $A$, $B$, and $q$ can make the identity hold on an interval.  The Kaniadakis operator therefore lies outside the Tsallis--entmax equivalence class, except at the common Shannon limit $\kappa\to0$.
For finite $\kappa$ the same two-power function is not affine in $\ln x$, so it is not a temperature-reparameterized Softmax law either.

The boundary gradient diverges, so all finite scores receive positive weight.  The algebraic tail follows directly from the exact root.  Because the entropy is even in $\kappa$, take $0<\kappa<1$ without loss of generality and define
\begin{equation}
 C_\kappa=\left(\frac{1-\kappa}{2\kappa}\right)^{1/\kappa}.
\end{equation}
For a low-score component with $x_i=(z_i-\theta)/T\to-\infty$,
\begin{align}
 \rho_i^{(\kappa)}
 &=C_\kappa(-x_i)^{-1/\kappa}
 \left[1+O(x_i^{-2})\right],
 \label{eq:ktail}\\
 \left.\frac{\partial\rho_i^{(\kappa)}}{\partial z_i}
 \right|_\theta
 &=\frac{C_\kappa}{\kappa T}(-x_i)^{-1-1/\kappa}
 \left[1+O(x_i^{-2})\right].
 \label{eq:kgradient-tail}
\end{align}
The exact fixed-$\theta$ derivative is
\begin{equation}
 d_i\equiv
 \left.\frac{\partial\rho_i^{(\kappa)}}{\partial z_i}\right|_\theta
 =\frac{\rho_i^{(\kappa)}}
 {T\sqrt{1-\kappa^2+\kappa^2x_i^2}}>0.
 \label{eq:kcomponent-derivative}
\end{equation}
It gives Eq.~\eqref{eq:kgradient-tail} directly.  Implicit differentiation of $\sum_i\rho_i^{(\kappa)}=1$ then gives the exact normalized Jacobian
\begin{equation}
 \frac{\partial\rho_i^{(\kappa)}}{\partial z_j}
 =d_i\left(\delta_{ij}-\frac{d_j}{\sum_\ell d_\ell}\right).
 \label{eq:kjacobian}
\end{equation}
Provided at least one other component remains away from the boundary, the diagonal response of a low-score component has the same leading power as Eq.~\eqref{eq:kgradient-tail}.  Thus both the weight and its normalized local sensitivity are suppressed by powers rather than exponentially.  The map remains dense unlike entmax, but Eqs.~\eqref{eq:ktail}--\eqref{eq:kjacobian} do not imply a nonzero entropy floor.  The relation $\exp_\kappa(x)\exp_\kappa(-x)=1$ expresses the reciprocal symmetry of the conventional deformation, although normalized attention probabilities themselves are not pairwise reciprocal.

\section{Global-moment projections and a boundary-gradient case}
\label{sec:projections}

\subsection{R\'enyi entropy: a distinct functional but not a distinct weight family}

R\'enyi entropy~\cite{renyi1961} can be written as
\begin{equation}
 H_q(\bm\rho)=\frac{1}{1-q}\ln M_q,
 \qquad M_q=\sum_i\rho_i^q,
 \label{eq:renyi}
\end{equation}
whereas Tsallis entropy in Eq.~\eqref{eq:tsallis-entropy} is $(1-M_q)/(q-1)$.  Their gradients satisfy the exact relation
\begin{equation}
 \nabla H_q=\frac{1}{M_q}\nabla S_q.
 \label{eq:renyi-gradient}
\end{equation}
Thus Eq.~\eqref{eq:equivalence} holds with $a=1/M_q>0$.  Because $M_q$ depends on the output profile, this is profile-shape equivalence, not fixed-parameter equivalence.  At fixed external $T$, the R\'enyi stationary equation contains the self-consistent global moment,
\begin{equation}
 E_i+\mu-T\frac{q}{1-q}
 \frac{\rho_i^{q-1}}{M_q}=0.
 \label{eq:renyi-stationary}
\end{equation}
Defining $T_{\rm eff}=T/M_q$ reduces it to the Tsallis stationary equation.  As $T$ varies, R\'enyi and Tsallis therefore trace the same set of normalized profile shapes.  At a prescribed $T$, however, R\'enyi attention is a self-consistent, input-adaptive-temperature map and is not identical to fixed-$T$ entmax.

The word \emph{stationary} is essential here.  R\'enyi entropy is concave on the full probability simplex for $0<q\leq1$, but it is not generally concave for $q>1$.  Its Hessian is
\begin{equation}
 \frac{\partial^2H_q}{\partial\rho_i\partial\rho_j}
 =-\frac{q}{M_q}\rho_i^{q-2}\delta_{ij}
 +\frac{q^2}{(q-1)M_q^2}
 \rho_i^{q-1}\rho_j^{q-1}.
 \label{eq:renyi-hessian}
\end{equation}
For example, at $q=2$ and $\bm\rho=(0.98,0.01,0.01)$, its value along the tangent direction $\bm d=(1,-1/2,-1/2)$ is
\begin{equation}
 \frac{\bm d^{\mathsf T}(\nabla^2H_2)\bm d}
 {\bm d^{\mathsf T}\bm d}\simeq0.637>0,
 \label{eq:renyi-nonconcave}
\end{equation}
which explicitly rules out concavity.  In this range, which includes the usual sparse entmax deformation, Eq.~\eqref{eq:renyi-stationary} still has the Tsallis componentwise form, but the variational objective need not be convex and a stationary branch need not be the unique global optimizer.  The gradient projection establishes operator-form equivalence; it does not restore concavity that the R\'enyi functional lacks.

This distinction corrects a tempting but misleading classification based only on whether the derivative contains a global moment.  Global coupling can make the temperature self-consistent while leaving the componentwise inverse-gradient law unchanged.  In that case the realization is implicit in its scale but not new in its normalized shape.

\subsection{Sharma--Mittal entropy and two-to-one-parameter projection}

The Sharma--Mittal entropy~\cite{sharma1975} is
\begin{equation}
 S_{q,r}^{\rm SM}(\bm\rho)
 =\frac{M_q^{(1-r)/(1-q)}-1}{1-r}.
 \label{eq:sm-entropy}
\end{equation}
The limit $r\to1$ gives R\'enyi entropy, $r\to q$ gives Tsallis entropy, and $q,r\to1$ gives Shannon entropy.  Differentiation yields
\begin{equation}
 \nabla S_{q,r}^{\rm SM}
 =M_q^{(q-r)/(1-q)}\nabla S_q.
 \label{eq:sm-gradient}
\end{equation}
The prefactor is positive in the interior of the simplex and is common to all components, but it depends on the output through $M_q$.  The stationary equation is therefore the Tsallis equation evaluated at
\begin{equation}
 T_{\rm eff}=T M_q^{(q-r)/(1-q)},
 \qquad
 \beta_{\rm eff}=\beta M_q^{(r-q)/(1-q)}.
 \label{eq:sm-temperature}
\end{equation}
Consequently,
\begin{equation}
 \rho_i^{\rm SM}
 =\left[\frac{q-1}{qT_{\rm eff}}
 (z_i-\theta)\right]_+^{1/(q-1)},
 \label{eq:sm-map}
\end{equation}
with $\theta$ fixed by normalization.  The two-parameter entropy family projects onto the one-parameter Tsallis profile shape.  The parameter $r$ remains present in the relation between the externally specified $T$ and $T_{\rm eff}$ and in quantities that depend on the value of the entropy functional, but it does not supply a second independent inverse-gradient shape.  At fixed external $T$, Eq.~\eqref{eq:sm-temperature} again defines a self-consistent map rather than ordinary entmax at a preassigned temperature.
As in the R\'enyi case, this is a statement about stationary form; whether that stationary point is the unique optimizer depends separately on the concavity domain in $(q,r)$.

Equations~\eqref{eq:renyi-gradient} and \eqref{eq:sm-gradient} expose a common mechanism.  If an entropy has the form $\widetilde S=F(S)$ with $F'(S)>0$, then
\begin{equation}
 \Pi_\Delta\nabla\widetilde S
 =F'(S)\Pi_\Delta\nabla S,
 \label{eq:monotone-projection}
\end{equation}
and the two entropies share a stationary profile shape after a generally state-dependent temperature reparameterization.  Generalizing an entropy by a monotone outer function can therefore increase the entropy parameter space without increasing the space of inverse-gradient shapes.  This optimizer equivalence is consistent with the recent projective maximum-entropy result for monotone transforms of a common power functional~\cite{hino2026}; the distinction made here is that a state-dependent prefactor still changes the score-to-weight map when the external temperature is held fixed.

\subsection{Burg entropy as a reciprocal-root boundary case}

Burg entropy~\cite{burg1975},
\begin{equation}
 S_{\rm B}(\bm\rho)=\sum_i\ln\rho_i,
 \label{eq:burg-entropy}
\end{equation}
is local and strictly concave in the simplex interior.  Its gradient diverges at the boundary, and the stationary equation gives
\begin{equation}
 \rho_i^{\rm B}=\frac{T}{E_i+\mu},
 \qquad
 \sum_i\frac{T}{E_i+\mu}=1,
 \label{eq:burg-map}
\end{equation}
where the admissible root obeys $E_i+\mu>0$ for every $i$.  Each component is explicit once $\mu$ is known, and normalization requires one scalar root.  Burg therefore has essentially the same realization structure as the Kaniadakis operator: componentwise inversion followed by scalar normalization.  Its distinguishing properties are instead the inverse-linear form and the boundary-divergent gradient.  We treat it as a boundary-gradient regime within local entropy-generated operators, not as an independent universal class or as evidence for a particular machine-learning application.

\section{Reciprocal-symmetric Abe attention}
\label{sec:abe}

\subsection{Entropy and classical stationarity}

The Jackson $q$-derivative provides one route to Tsallis entropy~\cite{johal1998}.  Abe symmetrized the construction under $q\leftrightarrow q^{-1}$ and obtained~\cite{abe1997}
\begin{equation}
S_{\Abe}(\bm\rho;q)
 =-\sum_i\frac{\rho_i^q-\rho_i^{1/q}}{q-q^{-1}},
 \qquad q>0,\quad q\ne1.
 \label{eq:abeentropy}
\end{equation}
It satisfies
\begin{equation}
 S_{\Abe}(\bm\rho;q)=S_{\Abe}(\bm\rho;q^{-1}),
 \qquad
 \lim_{q\to1}S_{\Abe}=S_{\rm Sh}.
 \label{eq:abesymmetry}
\end{equation}
The derivative is
\begin{equation}
 \frac{\partial S_{\Abe}}{\partial\rho_i}
 =-\frac{q\rho_i^{q-1}-q^{-1}\rho_i^{1/q-1}}
 {q-q^{-1}}.
 \label{eq:abederivative}
\end{equation}
Hence the stationary equation is
\begin{equation}
 \frac{q\rho_i^{q-1}-q^{-1}\rho_i^{1/q-1}}
 {q-q^{-1}}
 =\frac{z_i-\theta}{T}.
 \label{eq:abestationary}
\end{equation}

For $q>0$, reciprocal symmetry permits restriction to $q>1$.  Writing the left-hand side of Eq.~\eqref{eq:abestationary} as $G_q(\rho)$, one finds
\begin{equation}
 G_q'(\rho)=
 \frac{q(q-1)\rho^{q-2}
 -q^{-1}(q^{-1}-1)\rho^{q^{-1}-2}}
 {q-q^{-1}}>0
 \label{eq:abe-monotonicity}
\end{equation}
for $q>1$, and the $0<q<1$ result follows by $q\leftrightarrow q^{-1}$.  Thus $S_{\Abe}$ is strictly concave and the component inverse is monotone.  Its boundary derivative diverges, so every finite score has positive weight.  For fixed $\theta$, the sum of the component roots decreases continuously from infinity to zero as $\theta$ increases; the normalization root exists and is unique.  Abe attention is therefore an implicit but well-defined classical operator, although the two generally incommensurate powers in Eq.~\eqref{eq:abestationary} preclude a useful elementary inverse.

The exact backward map requires only the same implicit derivatives.  With
\begin{equation}
 c_i=\left.\frac{\partial\rho_i}{\partial z_i}\right|_\theta
 =\frac{1}{T G_q'(\rho_i)}>0,
\end{equation}
normalization gives
\begin{equation}
 \frac{\partial\rho_i}{\partial z_j}
 =c_i\left(\delta_{ij}-\frac{c_j}{\sum_\ell c_\ell}\right).
 \label{eq:abe-jacobian}
\end{equation}
Hence exact forward evaluation uses component and normalization roots, and the exact Jacobian is available without differentiating through an iterative solver.

The same functional-independence test used in Eq.~\eqref{eq:kappa-escape} separates Abe attention from the Tsallis family.  Away from $q=1$, Eq.~\eqref{eq:abederivative} contains the two distinct powers $\rho_i^{q-1}$ and $\rho_i^{1/q-1}$ with nonzero coefficients.  Their combination cannot equal a single Tsallis power plus a component-independent constant on an interval.  Reciprocal Abe deformation therefore changes the tangent-gradient direction rather than merely rescaling it.

\begin{figure}[!t]
 \includegraphics[width=\columnwidth]{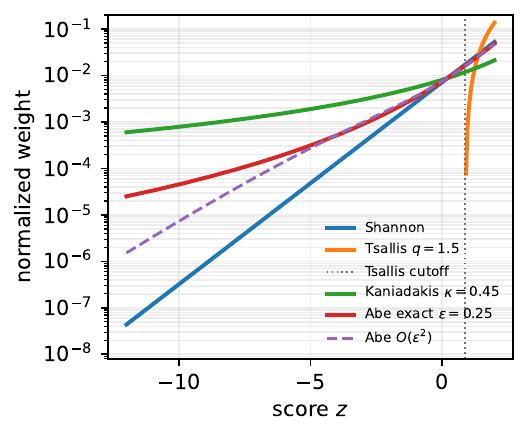}
 \caption{Normalized stationary weights on a common score grid.  Shannon attention decays exponentially; Tsallis attention ($q=1.5$) is identically zero to the left of the marked cutoff and is therefore absent there on the logarithmic axis; Kaniadakis attention ($\kappa=0.45$) has full support and a broad algebraic tail.  Exact classical Abe attention is shown for $q=e^{0.25}$ together with its normalized $O(\epsilon^2)$ approximation.  Parameters illustrate structural differences and are not performance optimized.}
 \label{fig:profiles}
\end{figure}

\subsection{Even-order selection rule}

Set
\begin{equation}
 q=e^\epsilon,
 \label{eq:epsilon}
\end{equation}
so that $q\leftrightarrow q^{-1}$ becomes $\epsilon\leftrightarrow-\epsilon$.  Let $\mathcal G_i(\rho_i,\mu,\epsilon)=0$ denote Eq.~\eqref{eq:abestationary} together with normalization.  The full system is even in $\epsilon$.  At $\epsilon=0$ its solution is the strictly positive Softmax vector $\bm p$, and the constrained Hessian of the Shannon objective is nonsingular on the tangent space.  The implicit-function theorem therefore gives a locally unique analytic branch $(\bm\rho(\epsilon),\mu(\epsilon))$.  Evenness and uniqueness imply
\begin{equation}
 \bm\rho(\epsilon)=\bm\rho(-\epsilon),
 \qquad \mu(\epsilon)=\mu(-\epsilon),
\end{equation}
and consequently
\begin{equation}
 \boxed{\bm\rho_{2n+1}=0,\quad \mu_{2n+1}=0}
 \qquad(n=0,1,\ldots).
 \label{eq:selection}
\end{equation}
All odd corrections vanish.  This is a symmetry selection rule for the normalized attention distribution, not merely a cancellation in a componentwise unnormalized kernel.

The symmetry also fixes the natural deformation coordinate for a learnable operator.  With $\lambda=\epsilon^2\geq0$, Eq.~\eqref{eq:abeexpansion} below begins linearly in $\lambda$, whereas $\partial\bm\rho/\partial\epsilon=0$ at $\epsilon=0$.  Consequently, optimizing $\epsilon$ after initializing it exactly at zero cannot leave the Softmax point through a first-order gradient update; one should instead treat a nonnegative $\lambda$ as the primary parameter or initialize $\epsilon$ away from zero.  The signs of the component corrections depend on the score profile, so reciprocal symmetry by itself does not guarantee that the largest and smallest weights increase simultaneously.

\subsection{Normalized perturbation through fourth order}

Define the Softmax reference distribution
\begin{equation}
 p_i=\frac{e^{-E_i/T}}{\sum_j e^{-E_j/T}},
 \qquad y_i=-\ln p_i,
 \label{eq:ydef}
\end{equation}
and let $\langle a\rangle_0=\sum_i p_i a_i$.  At fixed normalization multiplier, direct expansion of Eq.~\eqref{eq:abestationary} gives
\begin{align}
 f(y)&=y-y^2+\frac{y^3}{6},
 \label{eq:f}\\
 g(y)&=\frac{y}{360}\bigl(5y^5-87y^4+495y^3
 \notag\\[-2pt]
 &\hspace{2.7em}{}-1135y^2+1050y-330\bigr).
 \label{eq:g}
\end{align}
However, using $p_i[1+\epsilon^2f(y_i)+\cdots]$ directly violates normalization in general.  The multiplier must be expanded simultaneously.  Writing $a=\langle f\rangle_0$, the correctly normalized solution is
\begin{equation}
 \rho_i(\epsilon)=p_i\left[1+\epsilon^2c_{2i}
 +\epsilon^4c_{4i}+O(\epsilon^6)\right],
 \label{eq:abeexpansion}
\end{equation}
with
\begin{align}
 c_{2i}&=f(y_i)-\langle f\rangle_0,
 \label{eq:c2}\\
 c_{4i}&=g(y_i)-\langle g\rangle_0
 +a\left\{f'(y_i)-f(y_i)
 -\left\langle f'-f\right\rangle_0\right\}.
 \label{eq:c4}
\end{align}
Equations~\eqref{eq:c2} and \eqref{eq:c4} obey $\sum_i p_ic_{2i}=\sum_i p_ic_{4i}=0$ identically.  They are therefore tangent perturbations of the simplex.  The common subtractions are not optional renormalizations appended afterward; they arise from the $\epsilon$-dependence of the Lagrange multiplier.

The fourth-order coefficient is not generally $c_{2i}^2/2$ plus a common normalization term.  Thus the deformation is not a pure exponential of a quadratic-order correction.  More importantly, the polynomial truncation is local in $\epsilon$: for a broad score range it can produce negative approximate weights even while the exact implicit solution remains positive.  The perturbative form should be used only when $|\epsilon|$ is small enough that all retained probabilities remain nonnegative.  Exact Abe attention is obtained by solving Eq.~\eqref{eq:abestationary}, not by clipping the perturbative polynomial.

\section{Discussion}
\label{sec:synthesis}

\subsection{Operator families and thermodynamic identifiability}

Table~\ref{tab:comparison} separates stationary profile shape, fixed-$T$ scale, and realization complexity.  R\'enyi and Sharma--Mittal have the Tsallis inverse-gradient shape, but at a prescribed external $T$ their global moments produce self-consistent maps rather than ordinary fixed-temperature entmax.  Kaniadakis and Burg require a scalar normalization root after componentwise inversion but have different inverse-gradient laws.  Abe requires numerical component inversion together with normalization, while its small-$\epsilon$ branch is explicit once Softmax moments are evaluated.

\begin{table*}[t]
\caption{Structural properties of entropy-generated stationary maps under the linear score constraint.  ``Tsallis-shaped'' means equality of the componentwise inverse-gradient shape after a generally profile-dependent temperature rescaling; it does not mean equality of the fixed-$T$ maps, entropy functionals, or convexity.  R\'enyi entropy is not generally concave for $q>1$.}
\label{tab:comparison}
\begin{ruledtabular}
\small
\setlength{\tabcolsep}{4pt}
\resizebox{\textwidth}{!}{%
\begin{tabular}{llllll}
Entropy & Stationary profile & Fixed-$T$ scale & Support & Tail/cutoff & Realization\\
\hline
Shannon & Softmax & $T$ & full & exponential & closed-form sum\\
Tsallis ($q>1$) & entmax & $T$ & compact & exact cutoff & threshold/active set\\
R\'enyi & Tsallis-shaped & $T/M_q$ & branch dependent & as Tsallis & moment + threshold\\
Sharma--Mittal & Tsallis-shaped & Eq.~\eqref{eq:sm-temperature} & branch dependent & as Tsallis & moment + threshold\\
Kaniadakis & distinct root & $T$ & full & algebraic & scalar root\\
Burg & reciprocal root & $T$ & full & inverse linear & scalar root\\
Abe & reciprocal-symmetric & $T$ & full & deformed & component + scalar roots\\
\end{tabular}
}
\end{ruledtabular}
\end{table*}

This comparison makes the non-injectivity explicit.  The variational map sends an entropy to a stationary profile-shape class,
\begin{equation}
 S\longmapsto[\mathcal P_S],
 \qquad
 S_1\sim S_2
 \ \text{if Eq.~\eqref{eq:equivalence} holds}.
 \label{eq:quotient-map}
\end{equation}
It therefore acts on entropy functionals only after quotienting out normalization gauge and positive conformal rescaling of the tangent gradient.  R\'enyi and the complete Sharma--Mittal $r$ direction provide explicit nontrivial members of the Tsallis profile class.  The fixed-$T$ maps retain their moment-dependent scale.  Entropy space can therefore be higher-dimensional than the space of stationary profile shapes without making all corresponding implementations identical.

This quotient has a broader implication for generalized thermostatistics.  Within the linear-constraint setting, normalized stationary probabilities alone do not identify a unique generating entropy when the entropy scale is not calibrated independently.  A R\'enyi stationary profile can coincide with a Tsallis profile at a suitable effective temperature even though the functionals and their fixed-$T$ maps differ.  Distinguishing the underlying theories requires information not contained in a single stationary profile, such as the calibrated temperature relation, variational-potential value, second variations, fluctuation or response observables, or dynamics.  The present result does not assert that these quantities are equivalent; it identifies why stationary probabilities alone cannot settle that question.

\subsection{A practical test for new attention}

The criterion identifies what is needed to escape an existing profile class.  A deformation must change the \emph{direction} of the tangent gradient, not merely its common magnitude.  Kaniadakis entropy does so through the simultaneous powers $\rho_i^{\kappa}$ and $\rho_i^{-\kappa}$ in Eq.~\eqref{eq:kderivative}.  Abe entropy does so through reciprocal powers governed by $q\leftrightarrow q^{-1}$, with the additional consequence that its analytic branch is even in $\epsilon=\ln q$.  These structures cannot be removed even by a profile-dependent common temperature rescaling.

For model design, Eqs.~\eqref{eq:equivalence} and \eqref{eq:residual-test} turn this statement into an operational screen.  A proposed entropy should first be compared with known families after tangent projection.  A nonzero residual at one generic interior point disproves profile-shape equivalence immediately; residuals near zero at several points motivate, but do not replace, a symbolic check.  If the residual vanishes, one must then ask whether the proportionality factor is constant or output dependent.  Thus a new regularizer can define a distinct fixed-parameter map without adding a new inverse-gradient shape, and neither fact should be inferred from the number of entropy parameters alone.

The general construction of probability maps from entropy regularizers is established~\cite{niculae2017,blondel2019}; Softmax is Shannon regularized~\cite{bridle1990,shannon1948,jaynes1957}, and sparsemax and entmax already connect simplex optimization with Tsallis regularization~\cite{martins2016,peters2019}.  The contributions here are the attention-specific separation of profile-shape and fixed-parameter equivalence, its R\'enyi--Sharma--Mittal application, the exact Kaniadakis attention law and Jacobian with the trace-form factors retained, and the classical Abe operator and its normalized fourth-order hierarchy.  Recent uses of Tsallis--Hamiltonian training dynamics~\cite{lam2026} and interacting variational attention~\cite{wang2026via} address different variables.

\subsection{Selection, recoverability, and empirical scope}

The three principal tail structures imply different tradeoffs for machine learning.  Softmax exponentially suppresses a sufficiently low-scoring component; entmax removes it exactly after it crosses the active-set boundary; Kaniadakis attention retains both its weight and its componentwise sensitivity algebraically.  In this limited structural sense Kaniadakis attention preserves \emph{recoverability} of low-ranked candidates.  It does not decide which candidates are semantically relevant: that information must already enter through the query--key scores.

The same property has a negative side that rules out a blanket long-context claim.  An algebraic tail retains every sufficiently distant distractor more strongly than an exponential tail.  With many such distractors, their aggregate mass can therefore dilute the high-score set, while the compact support of entmax can be better suited to selective retrieval.  Kaniadakis attention should consequently be viewed as exchanging aggressive selection for continued low-score sensitivity, not as an intrinsic remedy for long-context attention.

Low attention entropy has been observed together with unstable Transformer training~\cite{zhai2023}.  Equations~\eqref{eq:ktail} and \eqref{eq:kgradient-tail} motivate testing whether algebraic rather than exponential suppression changes that training dynamics, but they do not establish such an effect.  Kaniadakis attention still approaches a concentrated distribution under sufficiently large score separation and has no nonzero entropy floor.  Likewise, the Abe parameter $\lambda=\epsilon^2$ supplies a reciprocal-symmetric shape coordinate, but the redistribution encoded by $c_{2i}$ is score dependent and is not a universal peak--tail improvement.

The analysis is intentionally restricted to structural statements following from the variational equations.  It does not claim that algebraic tails improve retrieval, that reciprocal symmetry improves robustness, or that any operator outperforms Softmax.  Scores are fixed during the optimization, masks must be imposed before normalization, and couplings among heads would require nonseparable entropy terms.  These limitations do not affect the stationary identities, but they delimit what can be inferred about a trained Transformer.

\section{Conclusion}

We have organized entropy-generated attention through the stationary gradient on the probability simplex.  Normalization gauge and positive conformal rescaling leave the inverse-gradient profile shape unchanged.  R\'enyi and Sharma--Mittal entropies consequently project onto the Tsallis shape despite their distinct functionals and, for Sharma--Mittal, its additional parameter.  Their output-dependent scale still makes the fixed-$T$ maps different from ordinary entmax.  More entropy parameters are therefore not synonymous with more independent attention-shape parameters.

Kaniadakis and Abe entropies lie outside the Tsallis profile class because they change the tangent-gradient direction.  Kaniadakis entropy yields an exact dense algebraic-tail root and an exact normalized Jacobian after one scalar normalization condition.  Classical Abe entropy yields a strictly concave reciprocal-symmetric map whose exact Jacobian follows from implicit differentiation and whose normalized expansion contains only even powers of $\epsilon=\ln q$.  Treating the multiplier consistently fixes the second- and fourth-order simplex-tangent corrections.  The resulting comparison separates profile shape, fixed-parameter scale, realization complexity, boundary behavior, and deformation symmetry without claiming an exhaustive universal classification.

The present work uses the Fisher-metric Lagrangian only through its stationary sector.  With $t$ interpreted as continuous layer depth, the next step is to test whether attention paths measured across sufficiently deep or weight-tied layers approximately follow the Euler--Lagrange dynamics, and whether the resulting relaxation and response quantities clarify machine-learning behavior.  That dynamical test is left for future study.

\appendix

\section{Normalization of the Abe expansion}
\label{app:abe-normalization}

This appendix records how the common multiplier shift produces Eqs.~\eqref{eq:c2} and \eqref{eq:c4}.  Before imposing normalization order by order, expansion of the component equation at a fixed multiplier gives
\begin{equation}
 \widetilde\rho_i
 =p_i\left[1+\epsilon^2f(y_i)
 +\epsilon^4g(y_i)+O(\epsilon^6)\right],
 \label{eq:unnormalized-abe}
\end{equation}
where $p_i=e^{-y_i}$, and $f$ and $g$ are given by Eqs.~\eqref{eq:f} and \eqref{eq:g}.  A change of the normalization multiplier appears as the same shift of the argument for every component.  We write it as
\begin{equation}
 y_i\longrightarrow y_i+\delta(\epsilon),
 \qquad
 \delta(\epsilon)=\epsilon^2a+\epsilon^4b+O(\epsilon^6).
 \label{eq:multiplier-shift}
\end{equation}
Expanding both the leading exponential and the functions of the shifted argument gives
\begin{align}
 \rho_i=p_i\bigg[1
 &+\epsilon^2\{f_i-a\}\notag\\
 &+\epsilon^4\left\{
 g_i+a(f_i'-f_i)+\frac{a^2}{2}-b
 \right\}+O(\epsilon^6)\bigg],
 \label{eq:shifted-expansion}
\end{align}
where $f_i=f(y_i)$ and $g_i=g(y_i)$.  The conditions $\sum_i\rho_i=1$ at second and fourth order determine the two common shifts,
\begin{align}
 a&=\langle f\rangle_0,
 \label{eq:a-shift}\\
 b&=\langle g\rangle_0
 +a\langle f'-f\rangle_0+\frac{a^2}{2}.
 \label{eq:b-shift}
\end{align}
Substitution into Eq.~\eqref{eq:shifted-expansion} gives
\begin{align}
 c_{2i}&=f_i-\langle f\rangle_0,\\
 c_{4i}&=g_i-\langle g\rangle_0
 +a\left[f_i'-f_i-\langle f'-f\rangle_0\right],
\end{align}
which reproduces Eqs.~\eqref{eq:c2} and \eqref{eq:c4}.  The construction makes $\sum_i p_i c_{2i}=\sum_i p_i c_{4i}=0$ automatic.  In particular, normalizing only after expanding each component would miss the derivative term $a(f_i'-f_i)$ and would give an incorrect fourth-order coefficient.

\begin{acknowledgments}
The author used ChatGPT (OpenAI, GPT-5.6 Sol) and Claude (Anthropic, Opus 5.0) to assist with scientific discussion, manuscript revision, and consistency checks. The author independently checked and verified all calculations, interpretations, numerical results, and conclusions and takes full responsibility for the content.
\end{acknowledgments}

\end{document}